\documentclass{article}
\usepackage[preprint]{neurips_2019}

\usepackage[utf8]{inputenc} 
\usepackage[T1]{fontenc}    
\usepackage{hyperref}       
\usepackage{url}            
\usepackage{booktabs}       
\usepackage{amsfonts}       
\usepackage{nicefrac}       
\usepackage{microtype}      
\usepackage{color}
\usepackage{amsmath}
\usepackage{graphicx}
\usepackage{wrapfig}
\usepackage{multirow}
\usepackage{amsmath,amssymb,amsfonts}
\usepackage[table,xcdraw]{xcolor}
\usepackage{threeparttable}

\title{DashNet: A Hybrid Artificial and Spiking Neural Network for High-speed Object Tracking}

\author{Zheyu Yang\textsuperscript{\rm 1}\thanks{These authors are equal contribution to this work.}, Yujie Wu\textsuperscript{\rm 1*}, Guanrui Wang\textsuperscript{\rm 1}, Yukuan Yang\textsuperscript{\rm 1}, Guoqi Li\textsuperscript{\rm 1}, \\ \textbf{Lei Deng\textsuperscript{\rm 1,3}, Jun Zhu\textsuperscript{\rm 2}, Luping Shi\textsuperscript{\rm 1}\thanks{Corresponding author: lpshi@mail.tsinghua.edu.cn}}\\ 
\\
\textsuperscript{\rm 1}\textbf{Center for Brain Inspired Computing Research (CBICR),}\\
  Department of Precision Instruments, \\
  Tsinghua University, 100084 Beijing, China\\
\\
\textsuperscript{\rm 2}\textbf{State Key Lab of Intelligence Technology and System,}\\
Department of Computer Science and Technology,\\
  Institute for AI, THBI Lab, \\
  Tsinghua University, 100084 Beijing, China\\
\\
\textsuperscript{\rm 3}\textbf{Department of Electrical and Computer Engineering,}\\
  University of California, Santa Barbara, CA 93106, USA
}

\begin{document}

\maketitle
	
\begin{abstract}

Computer-science-oriented artificial neural networks (ANNs) have achieved tremendous success in a variety of scenarios via powerful feature extraction and high-precision data operations. It is well known, however, that ANNs usually suffer from expensive processing resources and costs. In contrast, neuroscience-oriented spiking neural networks (SNNs) are promising for energy-efficient information processing benefit from the event-driven spike activities, whereas, they are yet be evidenced to achieve impressive effectiveness on real complicated tasks. How to combine the advantage of these two model families is an open question of great interest. Two significant challenges need to be addressed: (1) lack of benchmark datasets including both ANN-oriented (frames) and SNN-oriented (spikes) signal resources; (2) the difficulty in jointly processing the synchronous activation from ANNs and event-driven spikes from SNNs. In this work, we proposed a hybrid paradigm, named as \textbf{DashNet}, to demonstrate the advantages of combining ANNs and SNNs in a single model. A simulator and benchmark dataset “NFS-DAVIS” is built, and a temporal complementary filter (TCF) and attention module are designed to address the two mentioned challenges, respectively. In this way, it is shown that DashNet achieves the record-breaking speed of \textbf{2083FPS} on neuromorphic chips and the best tracking performance on NFS-DAVIS and PRED18 datasets. To the best of our knowledge, DashNet is the first framework that can integrate and process ANNs and SNNs in a hybrid paradigm, which provides a novel solution to achieve both effectiveness and efficiency for high-speed object tracking.

\end{abstract}

\section{Introduction}
	\begin{wrapfigure}{R}{0cm}
		\label{fig1}
  		\centering
			\includegraphics[width=6cm]{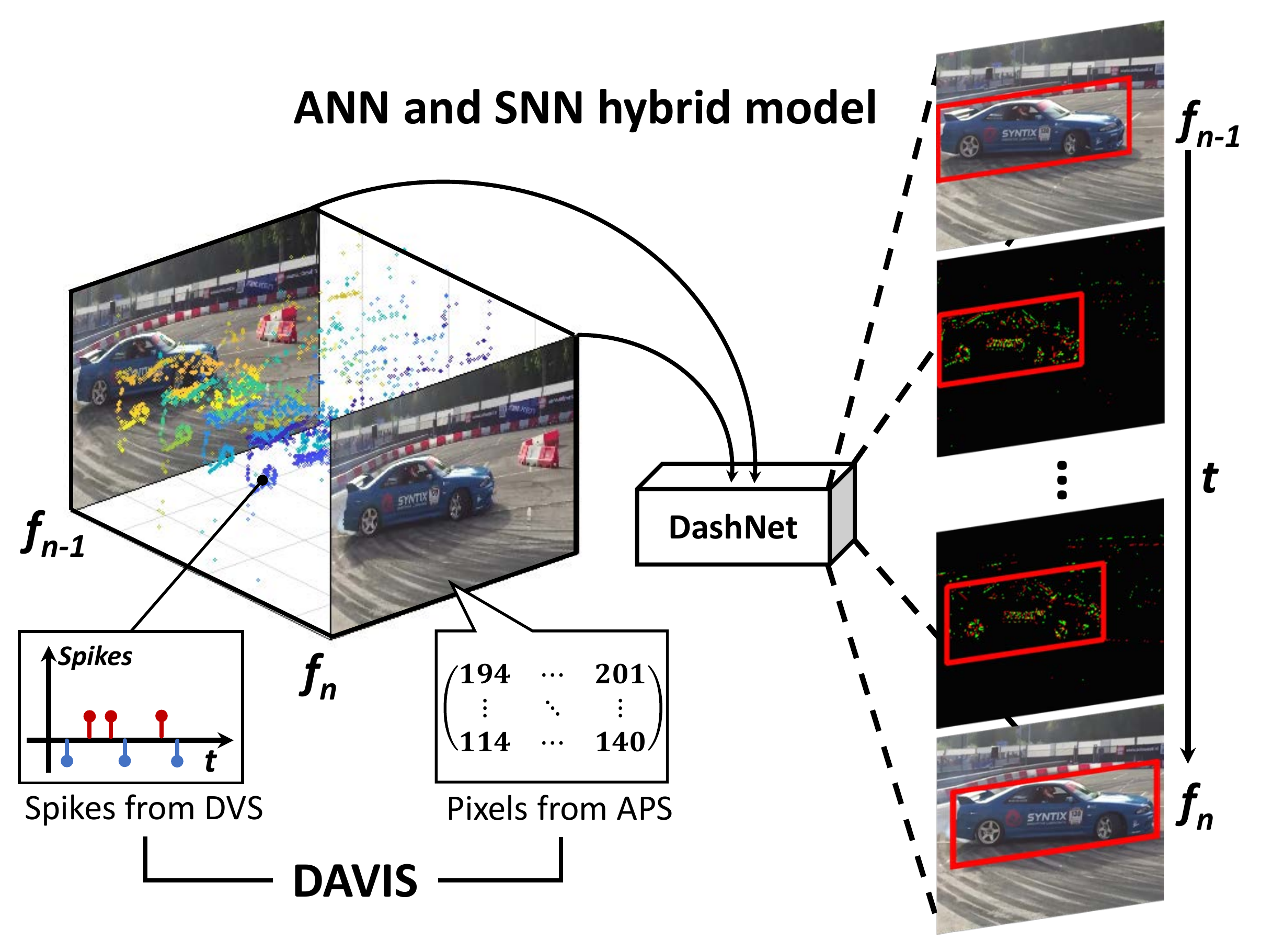}
  		\caption{\textbf{Overview of proposed hybrid approach}. DashNet can track the asynchronous spike-based data and synchronous frame-based data simultaneously (shown with red box).}
	\end{wrapfigure} 
    
    High-speed object tracking in complicated scenarios has been a long standing challenge in the computer vision communities. In recent years, emerging deep artificial neural networks (ANNs) have significantly promoted the development of this field via the capability of deep hierarchy and availability of the enormous amounts of data \cite{kristan2018sixth,kristan2017visual,Nature_Paper}. However, comparing with handcraft features methods, such as KCF \cite{henriques2015high} and Staple \cite{bertinetto2016staple}, those ANN-based methods have a great gap in speed, memory and computation costs owning to the dense matrix operation and high-precision information communication. Alternatively, benefit from the event-driven asynchronous processing and binary spike communication, brain-inspired SNNs are promising in ultra-low power implementation on neuromorphic hardwares \cite{merolla2014million,severa2018whetstone,Wu2018Spatio,wu2018direct}. Therefore, building a hybrid model to combine the efficiency and effectiveness of ANN trackers and SNN trackers is a promising direction for high-speed tracking. However, two challenges impede the development: (1) lacking the suitable benchmark to evaluate the hybrid model performance; (2) hard to jointly process the synchronous information from ANNs and event-driven spikes from SNNs.
    
    To address the first problem, we propose a hybrid data generation method. It can convert frame-based signals into a spike-based version by simulating the mechanism of event cameras \cite{Lichtsteiner2008A} and achieve higher temporal resolution by fully convolutional network (FCN) interpolation. In this way, the generated data, which similar to the signals of DAVIS (Dynamic and Active Pixel Vision Sensor) \cite{brandli2014a}, contains spike-based (DVS-based) and frame-based (APS-based) data simultaneously, as shown in Fig.\ref{fig1}. We use the proposed generation method to convert the NFS dataset \cite{kiani2017need} into a hybrid version (called NFS-DAVIS) for model evaluation.
    
    To solve the second problem, we propose a three-stage hybrid method: (1) the ANN or SNN is trained by the APS or DVS data, respectively; (2) a temporal complementary filter (TCF) is designed to merge the synchronous and asynchronous information from the ANN and SNN; (3) an attention module is involved, to 
    feed back the TCF outputs and provide more useful information for frame-based predictions.
    Furthermore, to evaluate the efficiency of our approach, we implement our method on both of general purpose computers(GPC) and neuromorphic-based platforms. Results demonstrate that our method achieves the best tracking performance and the record-breaking speed on neuromorphic chips. The main contributions of this work are summarized as follows:
    
    \begin{itemize}
        \item \textbf{A benchmark dataset for hybrid model }. We design a hybrid data simulator, which can generate the hybrid data from frame-based datasets. Through our simulator, we build the NFS-DAVIS$\footnote{The code and dataset will be released later.}$ dataset to favorably evaluate the performance of hybrid model.
        \item \textbf{A direct training approach for SNN tracking}. We develop a SNN model for event-driven based dataset and enabling accurate information representation for tracking. To our best knowledge, it is the first work towards exploring the direct training of SNNs for tracking and detection tasks.
        \item \textbf{A hybrid artificial and spiking model for high-speed tracking}. We propose a hybrid mode to jointly process the synchronous signals from ANNs and event-driven signals from SNNs via the TCF and attention mechanism. Subsequently, it achieves the best performance and the record-breaking 2083 FPS tracking speed on neuromorphic chips.
    \end{itemize}
    
\section{Related Work}
\begin{figure}
  \centering
	\includegraphics[width=0.8\linewidth]{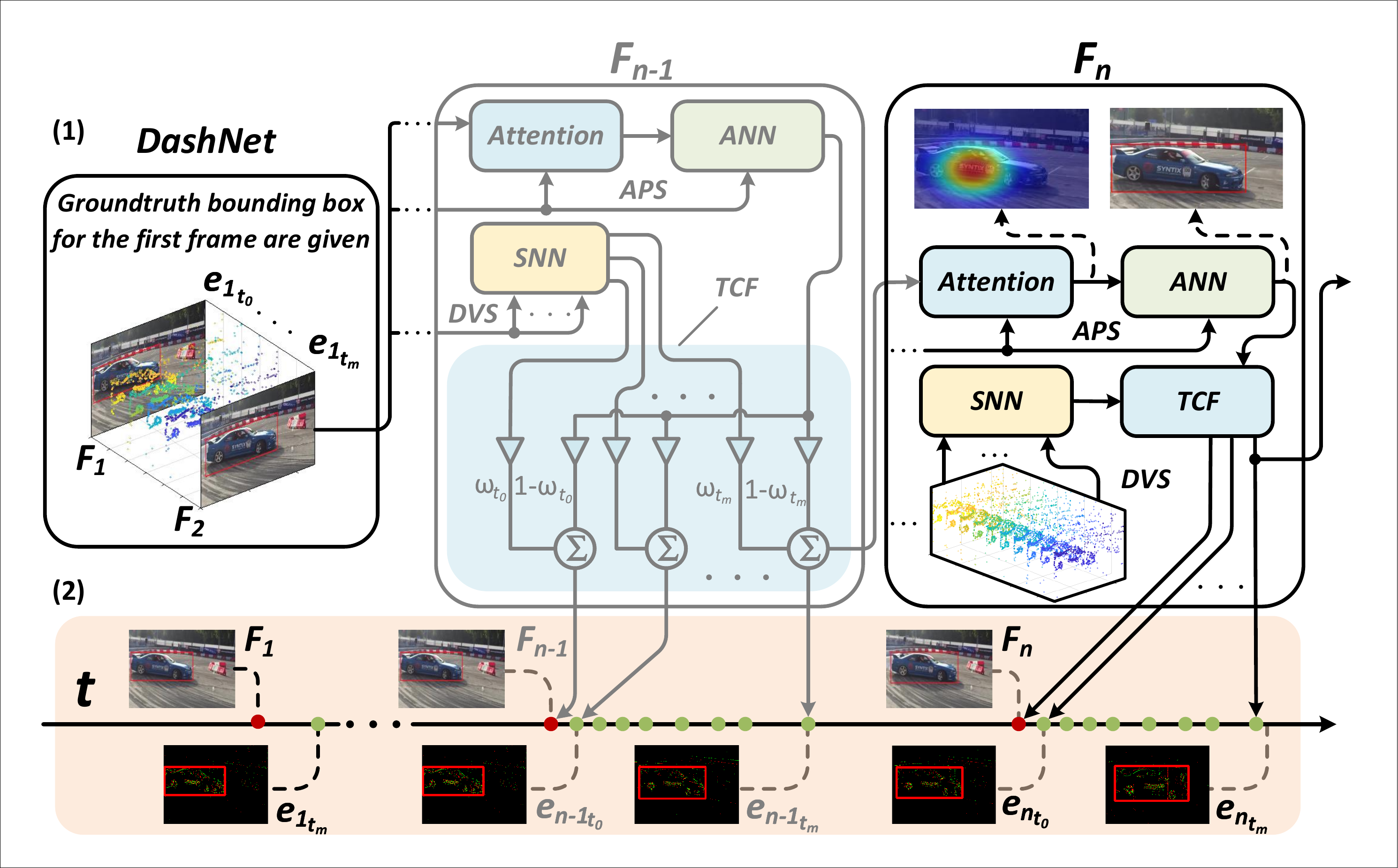}
  \caption{\textbf{Schematic of DashNet structure}. (1) The data flow between two adjacent frames $F_{n-1}$ (in the gray box) and $F_n$ (in the black box). Firstly, the ANN or SNN process the APS or DVS information respectively. Afterward, triggered by each SNN output, the TCF combines it with the nearest ANN output. Finally, when new $F_n$ coming, the attention module feedbacks the nearest TCF result to the ANN. (2) The timeline of the output sequence. the Green or red node represents the output timestamp of the SNN or ANN, respectively. }
  \label{fig2}
\end{figure}
We briefly review the works about the representative trackers, including CF-based as well as NN-based methods, and the related benchmark datasets and simulators. 

\textbf{Visual Tracking}. The mainstream of visual tracker can be divided into two types of methods: (1) CF-based methods, with cheap hand-crafted features (e.g. HOG \cite{dalal2005histograms}), have gained attention due to their computational efficiency and competitive performance. Starting from the MOSSE tracker \cite{bolme2010visual}, several variations have been investigated to improve the performance, such as kernelized correlation filters \cite{henriques2015high}, Staple \cite{bertinetto2016staple}, DSST \cite{danelljan2014accurate} etc. (2) Without any handcraft features, ANN-based methods tackle this task by training an end-to-end classifier online or offline, including MDNet \cite{nam2016learning}, FCNT \cite{wang2015visual}, GOTURN \cite{held2016learning} etc. Besides that, some works integrate the deep feature with CF-based tracker \cite{kiani2017learning, qi2016hedged, zhang2017multi}, which also can be regarded as variations of deep trackers. Generally, those methods achieve higher tracking performance but have a great gap in speed and computation costs with CF-based methods \cite{kristan2018sixth, kristan2017visual}. Therefore, our approach is focusing on how to improve the efficiency and effectiveness of deep trackers.

\textbf{Tracking Datasets and Simulators}. Standard tracking datasets, which capture annotated generic objects by the frame-based camera in real-world scenarios, have been widely used to evaluate tracking methods, such as VOT \cite{kristan2018sixth, kristan2017visual}, OTB100 \cite{wu2015object}, GOT-10k \cite{huang2018got} etc. In contrast, only a few works provide the benchmark dataset to evaluate the tracking performance of the hybrid model \cite{moeys2016steering, moeys2018pred18}. However, those datasets only capture a limited number of objects in a highly similar scene. This problem is mainly due to hybrid output sensors, such as DAVIS and Color-DAVIS\cite{li2015design, son20174}, are emerging direction in recent year and haven't been so common as the frame-based camera. As a compromise, some works propose the simulators of DAVIS based on the rendering engine and virtual scene modeling \cite{rebecq2018esim, mueggler2017event}. Those methods can reconstruct the hybrid output with high accuracy, but hard to generate a large amount of labeled data in numerous scenarios. In this paper, our simulator can directly generate the hybrid data from the labeled video dataset, avoiding the complex virtual scene modeling. 

\section{Method}
	\label{Method}
	In this section, we first develop an SNN model for tracking task in subsection.\ref{SNN_sec}, and briefly introduce the adopted ANN tracker in subsection.\ref{ANN_sec}. After that, we present two hybrid techniques (TCF and attention mechanism) for combining the results from single tracker. Finally, we propose a new data-augmentation method for  hybrid data generation.
\subsection{SNN Tracking}
\label{SNN_sec}
SNNs, as one of the brain-inspired networks, use the basic neural dynamics as its computational units.
As the name implies, neurons in SNNs communicate with each other in the form of an event-driven binary spike train, which is beneficial for low-energy computation of neuromorphic hardware systems, such as TrueNorth \cite{Akopyan2015TrueNorth}, Loihi \cite{Davies2018Loihi} and Tianjic \cite{Nature_Paper}. We adopt the discret Leaky Integrate-and-Fire(LIF) model as the basic computational
units in SNNs, shown as follows,
\begin{align}
 u^{n+1} = u^{n}e^\frac{-dt}{\tau}+\sum_i w_i^n o_i^{n-1},
    \label{lif_u}
\end{align}
where $dt$ denotes the discret timestep, $\tau$ denotes the time constant of membrane potential, $o_i^{n-1}$ denotes input spike train from the $n-1$th layer. Whenever membrane potential $u(t)^n$ exceeds a pre-defined threshold $u_{th}$, it will fire a spike $o(t)^n$ and reset to its resetting potential $u_{rest}$. We describe this process by the spike firing function $f_s$ mapping $u$ into $o$ as follows,
	\begin{align}
   		f_s(u) &= H(u(t) - V_{th})\\
   		u(t) &= u_{rest} \quad if \quad u(t) \geq V_{th},
   		\label{LIF}
	\end{align}
where $H(x)$ denotes the Heaviside function. It satisfies that $H(x)=1$ if $x >0$, otherwise $H(x)=0$.
As eq.\ref{LIF} illustrates, SNNs use the event-driven binary spike to represent information. However, tracking tasks usually require  a high-precision bounding box information around targets, which is different from the way of spike representation. Although rate coding can alleviate this problem, it needs a prohibitive time latency to guarantee performance \cite{Diehl2015Fast, Rueckauer2016Theory}.
To apply SNNs into the tracking tasks,  we regard the spiking network as a feature extractor and add a linear decoding layer in the last layer to learn high-precision information.
 In this way, it can accumulate the event-driven spike signals from previous layer over a very short time interval $[t, t+\delta t]$, and output final results through the linear-and-product operation. Accordingly, we propose the following loss function in which the mean square error for all samples under a given time windows $T$ is to be minimized
	\begin{align}
	\mathcal{L}_s = \sum_{i=1}^{n_N} ||\hat{o}_i- \sum_{j=1}^{n_{N-1}} w_{ij}^n\sum_{t \in [t_i, t_i +\delta t ]} o^{N-1}_j(t)  ||_2 
	+ \lambda \sum_{k=1}^{N}||W^{k}||_2,
	\label{loss_mse_snn}
	\end{align}
where $N$ denotes the number of network layer, $n_k$ denotes the neuron number in the layer $k$th,  $\lambda$ is a weight regularization parameter. 
If not specified, we use the upper index " $\hat{ }$ " to represent the ground truth in our paper. 
Given  eq.\ref{lif_u}-\ref{loss_mse_snn}, we can learn network parameters by updating gradients iteratively by back-propagation throughout time (BPTT)\cite{werbos1990backpropagation}. As for the non-differentialable points in firing function Eq.\ref{LIF}, we followed the work by \cite{Wu2018Spatio} to use a surrogate rectangle function for approximation.


\subsection{ANN Tracking}
\label{ANN_sec}
Neurons in ANNs communicate with each other using the high-precision and continuous activation. For the sake of completeness, we describe the information processing process of ANN as follows
\begin{align}
 a^n = f_a(\sum_{j} w_{i}^n a_i^{n-1}+b_j),
 \label{ANN_uints}
\end{align} 
where $a^n$ is the output neuronal activation, $a_j^{n-1}$ is the $j$-th afferent activation, $w_{ij}^{n}$ is the connection weight of synapse from neuron $j$ to the current neuron $i$ and $b_j$ is a bias term. The $f_a$ is a non-linear function, we use $f_a=RELU(x)=max(x,0)$ in our experiments. Unlike event-driven mechanism in SNNs, the basic operations described by eq.\ref{ANN_uints} usually are the expensive and high-dense matrix-vector addition and multiplication.

The input-output architecture of ANN is illustrated in Fig.\ref{fig2}. 
It receives the frame-based input generating by raw image and attention module, and output a four-dimensional vector $ a^N $ used for computing the bounding box. 
We train ANN module by back-propagation (BP) by minimizing the following square loss function $L_a$, in this case $n_N=4$, as follows:
\begin{align}
    	\mathcal{L}_a = \sum_{i=1}^{n_N} ||\hat{a}_i-  a^{N}_i  ||_2  
	+ \lambda \sum_{k=1}^{N}||W^{k}||_2,
	\label{loss_mse_ann}
\end{align}


\subsection{Hybrid Approach}
	\label{Hybrid module}

We introduce two hybrid blocks to integrate synchronous activations of ANN and asynchronous spikes of SNN. In subsection \ref{TCF_SC}, we present the temporal complementary filter that uses frame-based high-performance results to modify event-driven results. In subsection \ref{attention_SC}, we present a simple but effective attention mechanism that feedbacks event-driven results to provide more useful information for ANN recognition.

	\subsubsection{Temporal Complementary Filter (TCF)} 
	\label{TCF_SC}

    Given the tracking results of the ANN and SNN module, we introduce a temporal complementary filter to determine  the contribution of individual parts to the final result. More specifically, we let the output from the ANN or SNN be $\{a^N, o^N = [x, y, w, h] \}$, where $x$ and $y$ denote the center location of the objects, $w$ and $h$ denote the width and height of the objects, respectively. The temporal distance between the ANN output with timestamp $t_{a}$ and the SNN output with timestamp $t_{s}$ is defined by
           $D=||t_{a} - t_{s}||^2.$ 
 	For the identical $q$th output of SNN with timestamp $t_{s_q}$, we search the nearest $p$th output of ANN, in the set $P$ of all output result. The nearest distance satisfies
		\begin{align}
			D(t_{s_q}) = \min_{p \in P}||t_{a_p} - t_{s_q}||^2.
		\label{nearest neighbor temporal distance}
		\end{align}
		In this way, we can compute the combination weight $\omega$ respect to the SNN part by 
		\begin{align}
			\omega(D(t_{s_q})) = \frac{- 1 + \exp(D(t_{s_q}))}{1 + \exp(D(t_{s_q}))} = 1 - \frac{2}{1 + \exp(D(t_{s_q}))}.
		\label{temporal complementary filter weight}
		\end{align}
	Finally, we can obtain the combination output $h$ by
		\begin{align}
			h(t_{s_q}) = \omega (D(t_{s_q})) o(t_{s_q}) + (1 - \omega(D(t_{s_q})))a(t_{a_p}),
		\label{temporal complementary filter}
		\end{align}
	where $o$ and $a$ denote the output results from the SNN tracker and the ANN tracker, respectively, and we omit the upper index $N$ for simplicity. 
		
	The TCF assumes that ANN-based outputs can provide more accurate tracking results, which can be helpful for modifying SNN-based results periodically. Hence, eq.\ref{nearest neighbor temporal distance}-\ref{temporal complementary filter} describe that at time $t_{s_q}$, if there is a neighboring ANN output ($D \to 0$), we assign a small value of combination weight to SNN output ($\omega \to 0$) while enhancing the combination weight of ANN ($1-\omega$), otherwise vice versa.
 
	\subsubsection{Attention Mechanism} 
	\label{attention_SC}

We develop an attentional method as a trainable channel plugging in the input layer of standard CNN networks. The implementation can be divided into two phases:

\textbf{In inference phase}, when predicting for frame-based picture $F_t$, we first search the nearest TCF tracking trajectory and use the feedback results to generate a new input tensor $\hat{F_t}$.
More formally, we let the frame-based image be $F_t \in R^{W\times H\times C }$ and tracking trajectory be $h_t (t \in \{1,2,3,...,T\})$, where $ W, H, C $ refer  to the size of input images. When tracking the $\hat{t}$th frame objective, similar with eq.\ref{nearest neighbor temporal distance} we first search the nearest TCF output $h(t_k)$ from eq.\ref{temporal complementary filter} with $t_k$ timestamp,
\begin{align}
t_k = argmax_{t \in T}\{t| t<\hat{t}\}.
\end{align}
and use it to generate a soft mask weight $M^{\hat{t}}$. The element $M_{i,j}$ of mask matrix is defined by
\begin{align}
M^{t}_{i,j} =  \begin{cases}
1  & (i,j) \in B(o_{t_k})\\
\frac{1}{2\pi\sqrt{\sigma}} e^{-\frac{(i-x)^2+(j-y)^2}{2\sigma^2}}, \quad & (i,j) \not \in B(o_{t_k}),
\end{cases}
\end{align}
where the boxing box area $B(h(t_k))$ is determined by the tracking state $h(t_k)$. After that, we produce the attentional channel input $A^t$ as $A_t = X_t \star M^t$ and concatenates it with the original image into a new tensor $\hat{X} \in R^{W\times H\times 2C }$ as network input.

\textbf{In training phase},
we take a simple data-augmentation method to reduce the computational demanding for generating attention masks. We make a random shift $\beta$ and random scaling $\alpha$ on the ground true frame-based label $\hat{a}_t$ to replace the generated signal from TCF's tracking trajectory $h'(t_k)$
\begin{align}
\nonumber
	h'(t_k) &= (1 + \alpha)(\hat{a}_t + \beta) \\
	\alpha &\sim N (0, 0.1), \quad \beta \sim N(0, 0.1).
\end{align}
where we assume that $\alpha$ and $\beta $ obey the normal distribution $N(u,\sigma)$. In this way, the frame-based CNN can learn the noisy auxiliary information provided by the event-driven dataset, which helps to improve the robustness and reduce the communication.
	

	\begin{figure}
  		\centering
			\includegraphics[width=0.8\linewidth]{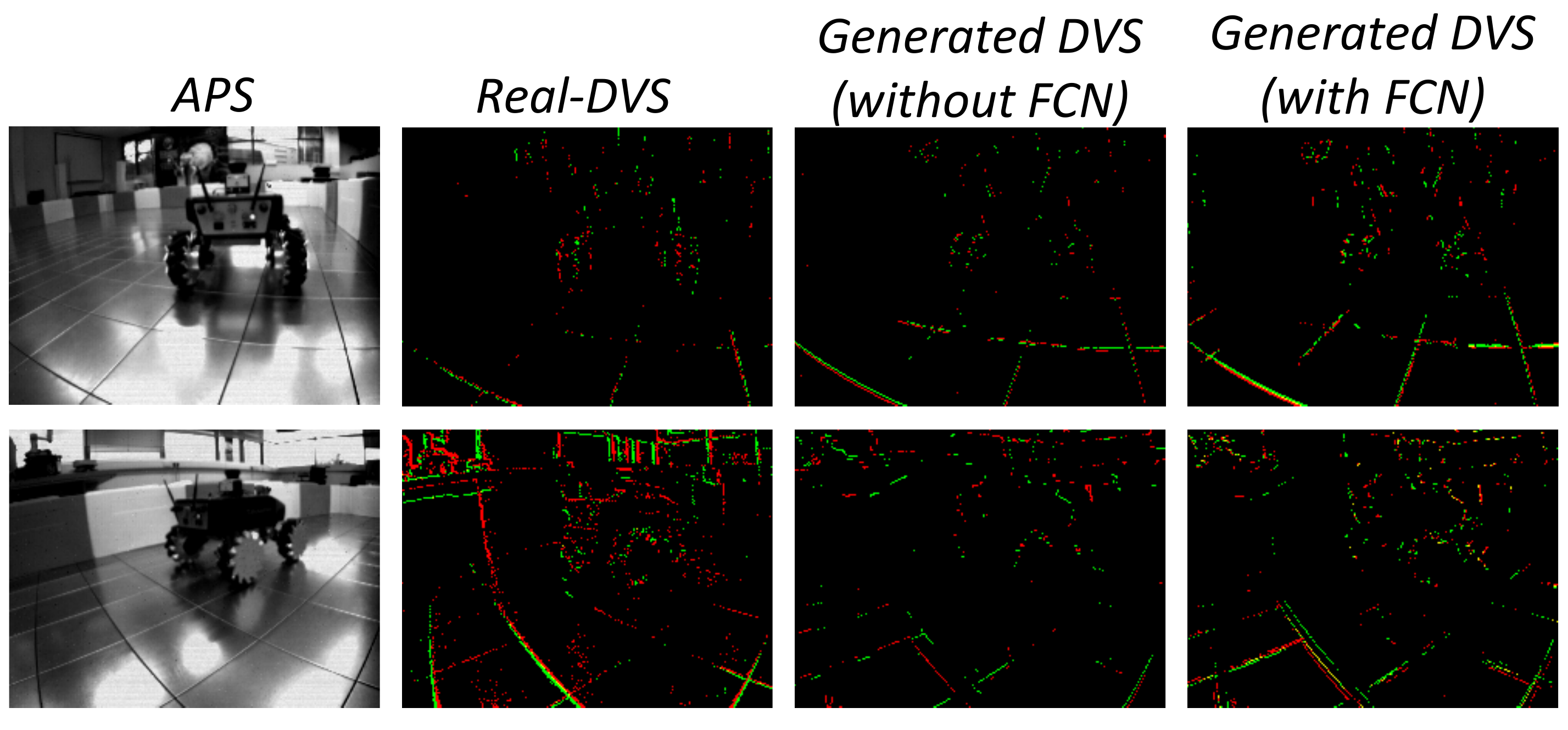}
  		\caption{\textbf{Side-by-side comparison of simulated data with real DAVIS data}. The first and second columns illustrate the real APS and DVS data captured by the DAVIS camera. The third and fourth columns illustrate the generated data with or without FCN interpolation. }
  		\label{fig3}
	\end{figure} 
	
\subsection{Hybrid Data Simulator}

We focus on generating a dataset containing frame-based and spike-based signals simultaneously at low costs. To this end, we first use the FCN to interpolate video for improving temporal resolution and then generate the APS or DVS stream by simulating the mechanism of DAVIS.
	
	\textbf{Video Interpolation}. Similar with \cite{niklaus2017video1, niklaus2017video2, niklaus2018context}, we use the FCN for video interpolation. During the inference, the network input with a pair of images $(I_{n-1},I_{n+1})$, which means the last and next frames. As the eq.\ref{FCN}, the patches $P_{n-1}(x, y)$ and $P_{n+1}(x, y)$ from $I_{n-1}$ and $I_{n+1}$ convolute with kernel $K$, and output an interpolated frame $I_n$,
	\begin{align}
	\label{FCN}
		I_n(x, y) = [P_{n-1}(x, y) \quad P_{n+1}(x, y)] * K(x, y).
	\end{align} 
	In the training phase, we randomly pick up the adjacent three frames from video sequences and extract the middle frame as the ground truth. Therefore, we choose the $\mathcal{L} _1$-norm based per-pixel difference loss $\mathcal{L}_1 = ||I_n - I_{gt}||_1$ as our loss function. 
	
	\textbf{Hybrid Data Modeling}. The DAVIS data jointly include asynchronous spikes from DVS and synchronous frames from APS. In the APS part, we directly extract non-interpolation frames in $t_f$ with a fixed temporal interval $\Delta t_f$ and contained $m$ frames. In the DVS part, we denote one DVS event by $\{x, y, p, t_e\}$, where $x$ and $y$ denote the pixel coordinates of the event, $p \in \{-1,+1\}$ denotes the polarity of the event, $t_e$ denotes the timestamp of the event. Note that, the DVS spike will fire, when the brightness changes higher than a pre-set threshold $\theta$ within the minimal temporal resolution $\Delta t_e$ \cite{Lichtsteiner2008A, brandli2014a}. We define brightness changes as $\Delta I(x, y) = I_{n}(x, y) - I_{n-1}(x, y)$, from the eq.\ref{FCN}. Then the $\hat{p}$ can be simulated by a unit STEP function $\epsilon$ as follows:
	\begin{align}
		p(x, y) &= \epsilon(\Delta I(x, y) - \theta) + \epsilon(\Delta I(x, y) + \theta) - 1.
	\end{align}
	When $p(x, y) \neq 0$, the firing rate $f_e$ , as same as the number of spikes between $I_{n}$ and $I_{n-1}$ can be calculate by $f_e(x, y) = \Delta I(x, y) / \theta$. Due to DVS has random temporal jitter between $I_{n}$ and $I_{n-1}$, we simulate the timestamp of each spike by the uniform distribution, $t_e(x, y) = t_f + \lfloor U(0, \frac{\Delta t_f}{m \cdot \Delta t_e}) \rfloor \cdot \Delta t_e$, where "$\lfloor$ $\rfloor$" denotes a floor function.
\section{Experiments and Results}
	\begin{figure}
  		\centering
			\includegraphics[width=1\linewidth]{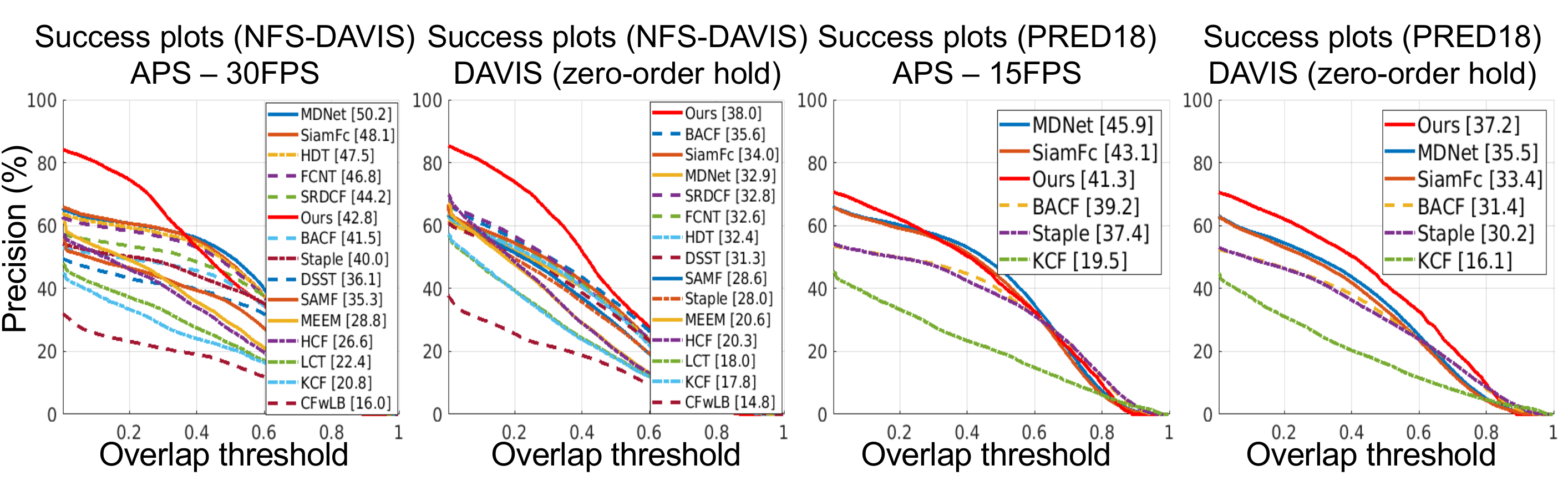}
  		\caption{Success plots of our approach and the state-of-the-art trackers on NFS-DAVIS and PRED18.}
  		\label{fig4}
	\end{figure}  

In this section, we evaluate the performance of DashNet from 
software simulation to hardware implementation on  NFS-DAVIS and PRED18 \cite{moeys2016steering, moeys2018pred18} datasets, comprehensively. The structure of ANN is  Input-32C3S3-MP2-64C3S1-MP2-128C3S1-128C1S1-MP2-256C2S2-FC1024-FC4, and the structure of SNN is Input-MP2-32C3S3-64C3S1-128C3S1-128C3S2-256C3S2-FC1024-FC4. We used ADAM optimizer to accelerate training. The more details about our experimental setup, simulator configuration, and dataset introduction are shown in the supplementary materials.
\subsection{Performance of Hybrid Data Simulator}
   To quantify the accuracy of our simulator, one possible solution is to directly compare the generated data with "ground truth" real-DAVIS data. However, the NFS doesn't have a real-DAVIS version. Instead, we generate the DAVIS data based on PRED18 dataset and estimate the root mean squared error (RMSE) and peak signal-to-noise ratio (PSNR) of simulated data with respect to a ground truth data obtained by oversampling the original signal with 15, 30, 60 and 120Hz. As Table.\ref{tab:my-table} shown, with the FCN interpolation the generated data achieve higher simulation performance and higher temporal resolution. To show qualitatively of our simulated DAVIS data resembles that of a real DAVIS camera, we show a side-by-side comparison of the real and simulated data in fig.\ref{fig3}.

\begin{table}[!htbp]
\centering
\begin{small}
\caption{Simulator performance comparison with or without FCN interpolation.}
\label{tab:my-table}
\begin{tabular}{l|c|c|c|c}
\hline
 & \begin{tabular}[c]{@{}c@{}}without\_interpolation\\ (15Hz)\end{tabular} & \begin{tabular}[c]{@{}c@{}}interpolation\_1$^*$\\ (30Hz)\end{tabular} & \begin{tabular}[c]{@{}c@{}}interpolation\_3\\ (60Hz)\end{tabular} & \begin{tabular}[c]{@{}c@{}}interpolation\_7\\ (120Hz)\end{tabular} \\ \hline
RMSE & 0.437$\pm$0.111 & 0.356$\pm$0.101 & 0.288$\pm$0.098 & 0.247$\pm$0.078 \\
PSNR (dB) & 55.81$\pm$2.40 & 57.58$\pm$2.87 & 59.45$\pm$2.96 & 60.48$\pm$2.79 \\ \hline
\end{tabular}
\begin{tablenotes}
\item[1] * interpolation\_"n": interpolating n frames in two adjacent frames.
\end{tablenotes}
\end{small}
\end{table}

\subsection{Ablation Studies}
   We investigate how the contribution of TCF and attention mechanism to our method performance. In the TCF part, we compare the performance of the separate SNN with the after-hybrid SNN. We find that with the ANN's assistance the after-hybrid SNN can achieve higher performance on both mean intersection over union (mIOU), mean area under the curve (mAUC), and robustness (Rb) \cite{wu2015object}. In the attention part, we compare the separate ANN with or without the attention module (ANN-AT and ANN-NAT). Afterwards, we compare the hybrid module performance against the ANN-AT outputs after zero-order hold interpolation. Therefore, we can prove that the hybrid approach can track features between the blind time of two frames and significantly improve the temporal resolution of the model. The overall comparisons are shown in Fig.\ref{fig4} and details are shown in Table.\ref{table1}. Through above comparisons, these results demonstrate that the TCF and attention module can significantly improve the performance of the tracker with both mIOU, mAUC, and Rb.
\begin{table}[!htbp]
\begin{small}
\centering
\caption{Comparisons of DVS-based, APS-based, and DAVIS-based performance with mIOU, mAUC (0.5 IOU) and Rb on NFS-DAVIS and PRED18 datasets.}
\label{table1}
\resizebox{\textwidth}{!}{%
\begin{tabular}{l|c|cc|cc|cc}
\hline
\multirow{2}{*}{Dataset}   & \multirow{2}{*}{Benchmark} & \multicolumn{2}{c|}{DVS}                   & \multicolumn{2}{c|}{APS}                   & \multicolumn{2}{c}{DAVIS}                                                                     \\ \cline{3-8} 
                           &                            & SNN-origin      & SNN-hybrid               & ANN-NAT         & ANN-AT                   & \begin{tabular}[c]{@{}c@{}}ANN-AT\\ (zero-order hold)\end{tabular} & Hybrid-AT                \\ \hline
\multirow{3}{*}{NFS-DAVIS} & mIOU                       & 0.364$\pm$0.034 & \textbf{0.380$\pm$0.025} & 0.401$\pm$0.023 & \textbf{0.412$\pm$0.020} & 0.363$\pm$0.021                                                    & \textbf{0.397$\pm$0.027} \\
                           & mAUC                       & 0.299$\pm$0.038 & \textbf{0.316$\pm$0.028} & 0.392$\pm$0.024 & \textbf{0.438$\pm$0.027} & 0.332$\pm$0.023                                                    & \textbf{0.383$\pm$0.025} \\
                           & Rb                         & 0.329$\pm$0.039 & \textbf{0.28$\pm$0.031}  & 0.218$\pm$0.029 & \textbf{0.206$\pm$0.024} & \textbackslash{}                                                   & \textbf{0.225$\pm$0.031} \\ \hline
\multirow{3}{*}{PRED18}    & mIOU                       & 0.306$\pm$0.040 & \textbf{0.326$\pm$0.036} & 0.353$\pm$0.024 & \textbf{0.375$\pm$0.025} & 0.273$\pm$0.030                                                    & \textbf{0.338$\pm$0.028} \\
                           & mAUC                       & 0.321$\pm$0.043 & \textbf{0.353$\pm$0.039} & 0.393$\pm$0.027 & \textbf{0.413$\pm$0.021} & 0.274$\pm$0.032                                                    & \textbf{0.372$\pm$0.031} \\
                           & Rb                         & 0.356$\pm$0.041 & \textbf{0.267$\pm$0.042} & 0.232$\pm$0.023 & \textbf{0.212$\pm$0.024} & \textbackslash{}                                                   & \textbf{0.244$\pm$0.028} \\ \hline
\end{tabular}%
}
\end{small}
\end{table}

\subsection{Overall Performance}
\begin{figure}
  \centering
	\includegraphics[width=1\linewidth]{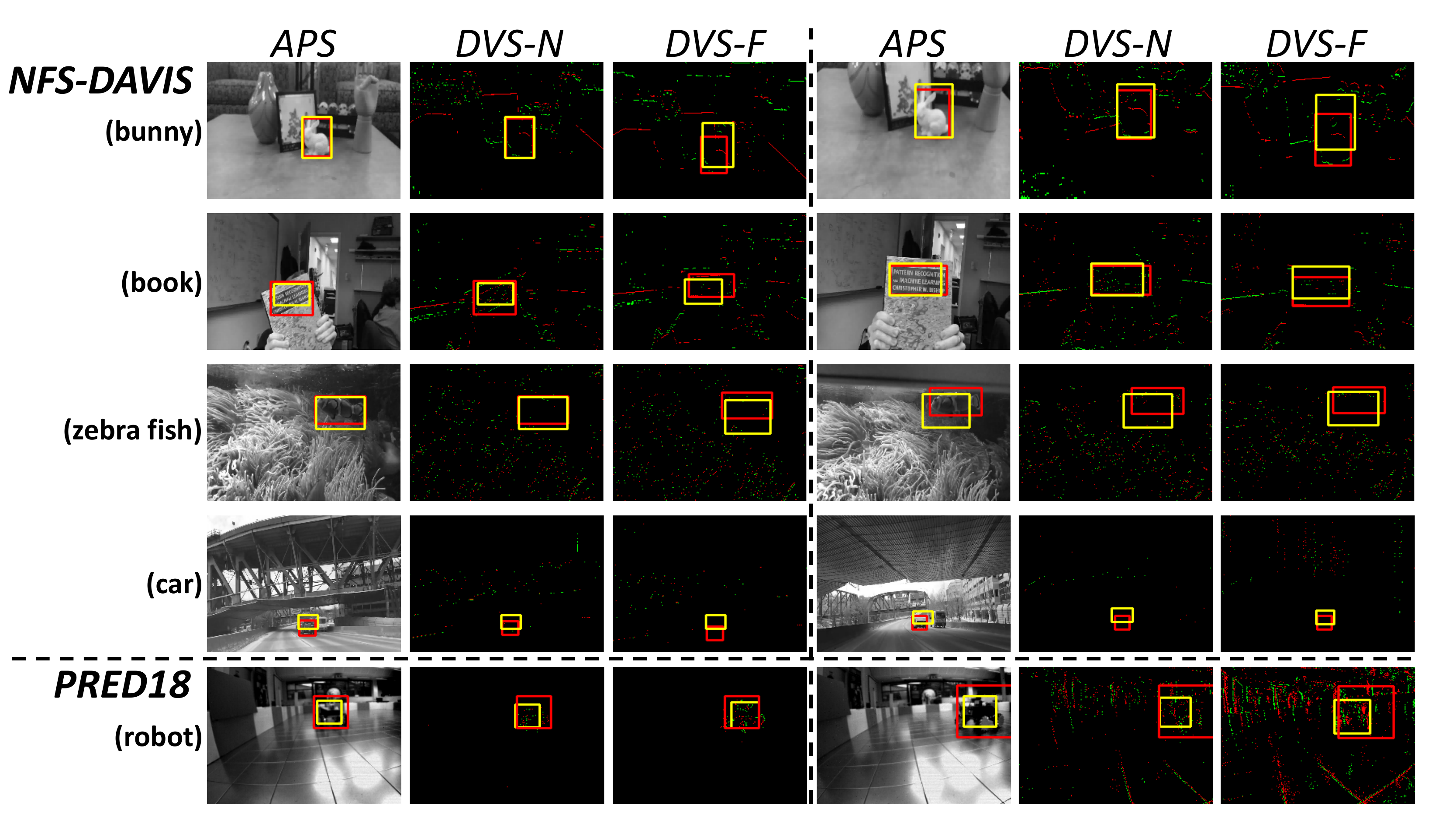}
  \caption{Qualitative results of our method. Bunny, book, car and zebrafish are from NFS-DAVIS's test set; the robot is from PRED18's test set \cite{moeys2016steering, moeys2018pred18}. The red box denotes the ground truth label, the yellow bounding box denotes results from our method.}
    \label{fig5}
\end{figure}

We compare the proposed model with three types of competitive trackers on PRED18 and NFS-DAVIS datasets, including: (1) CF-based trackers (DSST \cite{danelljan2014accurate}, SAMF \cite{li2014scale}, SRDCF \cite{danelljan2015learning}, KCF \cite{henriques2015high}, LCT \cite{ma2015long}, CFLB \cite{kiani2015correlation}, Staple \cite{bertinetto2016staple}, BACF \cite{kiani2017learning}, HCF \cite{ma2015hierarchical}, HDT \cite{qi2016hedged}); (2) ANN-based trackers (MDNet \cite{nam2016learning}, SiameseFC \cite{bertinetto2016fully-convolutional}, FCNT \cite{wang2015visual}); (3) SVM-based tracker (MEEM \cite{zhang2014meem}). We show the top-10 results of those contrast models for clarity. Notably, because such type of multimodal datasets is an emerging direction in recent year, we find that there is no suitable contrast model to compare performance on APS-based and DVS-based data simultaneously. 
As a compromise, we compensate the DAVIS-based results of other pure APS-based tracker using the zero-order hold interpolation, which keep same tracking trajectory in the blind time between two adjacent frames. In this way, we give a comprehensive comparison shown in the Table.\ref{table2}. It indicates that our method achieves the best performance in the high-speed DAVIS. Besides that, we simulate our model on a Tianjic \cite{Nature_Paper} multi-chip systems
and compare other tracker on general purpose computers  with CPUs of Intel Core i7, GPUs of Nvidia Titan X. It shows that our method has competitive speed with other trackers on GPC devices. Because of the challenges in algorithm design, to our best knowledge, no work has been reported to implement such SNN-based tracking task on neuromorphic hardware simulation. For the first time, Our work demonstrates the SNN-based tracking performance and shows the superiority of hybrid model for high-speed tracking tasks with ultra-high tracking speed of 2083.3 FPS. The more details about our method's hardware performance are shown in subsection.\ref{hardware}. The Fig.\ref{fig5} shows qualitative results of our approach on NFS-DAVIS and PRED18 test sets. The "DVS-N" or "DVS-F" means the nearest or the farthest SNN's outputs of the current frame "APS", respectively.
\begin{table}[!htbp]
\centering
\caption{Comparisons with the state-of-the-art trackers on NFS-DAVIS and PRED18 datasets. Results are reported as the AUC scores (\%) at 0.5 IOU and the speed of each tracker on CPUs, GPUs or neuromorphic chips.}
\label{table2}
\resizebox{\textwidth}{!}{%
\begin{tabular}{l|ccccccccccc}
\hline
Dataset            & Ours             & BACF             & SRDCF            & Staple           & DSST             & KCF              & SAMF             & HDT              & MDNet            & SiameseFC        & FCNT             \\ \hline
NFS-DAVIS          & \textbf{38.0}    & 35.6             & 32.8             & 28.0             & 31.3             & 17.8             & 28.6             & 32.4             & 32.9             & 34.0             & 32.6             \\
Pred18             & \textbf{37.2}    & 31.4             & \textbackslash{} & 30.2             & \textbackslash{} & 16.1             & \textbackslash{} & \textbackslash{} & 35.5             & 33.4             & \textbackslash{} \\ \hline
Speed-CPU          & \textbackslash{} & 38.3             & 3.8              & 50.8             & 12.5             & \textbf{170.4}   & 16.6             & 9.7              & 0.7              & 2.5              & 3.2              \\
Speed-GPU          & 47.8             & \textbackslash{} & \textbackslash{} & \textbackslash{} & \textbackslash{} & \textbackslash{} & \textbackslash{} & 43.1             & 2.6              & 48.2             & \textbf{51.8}    \\
Speed-Neuromorphic & \textbf{2083.3}  & \textbackslash{} & \textbackslash{} & \textbackslash{} & \textbackslash{} & \textbackslash{} & \textbackslash{} & \textbackslash{} & \textbackslash{} & \textbackslash{} & \textbackslash{} \\ \hline
\end{tabular}%
}
\end{table}

\subsection{Evaluation on Neuromorphic Hardware}
\label{hardware}


In order to facilitate practical applications, we implement the DashNet on neuromorphic chips. The results are recorded in Table.\ref{table3}. The details of hardware implementation are followed by works \cite{deng2018semimap, shi2015development}. As shown at the second and third column in Table.\ref{table3}, due to the binary spike features, the SNN bypasses multiplication operations and only has addition operations. Furthermore, through the event-driven and sparse feature, the SNN can also reduce the total operations and significantly alleviate the bandwidth pressure. Therefore, with above spike-based features, the SNN achieve a higher tracking speed (2083 FPS) and better power efficiency (127.9 FPS/W) than the ANN; Otherwise, the ANN has higher tracking accuracy (39.3\% and 40.7\% mAUC) but lower power efficiency than the SNN. DashNet makes a balance on efficiency and accuracy, which works often on SNN mode in real-world tracking scenarios. As Table.\ref{table3} shown, DashNet inherits the advantages of the ANN and SNN, leading to its simultaneous improvement in efficiency (2083 FPS speed and 100.1 FPS/W power efficiency) and effectiveness (37.2\% and 38.0\% mAUC).


\begin{table}[!htbp]
\centering
\begin{small}
\caption{Evaluation of DashNet performance based on neuromorphic chips.}
\label{table3}
\begin{tabular}{l|cc|cc|cc}
\hline
Model & \begin{tabular}[c]{@{}c@{}}\#Add.  \\ ($10^6 \times$Ops$^*$)\end{tabular} & \begin{tabular}[c]{@{}c@{}}\#Mul. \\ ($10^6 \times$Ops)\end{tabular} & \begin{tabular}[c]{@{}c@{}}Speed\\ (FPS)\end{tabular} & \begin{tabular}[c]{@{}c@{}}Power efficiency\\ (FPS/W)\end{tabular} & \begin{tabular}[c]{@{}c@{}}mAUC\\ (PRED18)\end{tabular} & \begin{tabular}[c]{@{}c@{}}mAUC\\ (NFS-DAVIS)\end{tabular} \\ \hline
ANN & 32.4 & 32.3 & 240.5 & 55.4 & 0.393 & 0.407 \\
SNN & 33.0 & 0.0 & 2083.3 & 127.9 & 0.321 & 0.302 \\
DashNet & 36.7 & 3.7 & 2083.3 & 100.1 & 0.372 & 0.380 \\ \hline
\end{tabular}
\begin{tablenotes}
\item[] * Ops: average number of operations in the inference.
\end{tablenotes}
\end{small}
\end{table}


\section{Conclusion}
\label{others}
In this paper, we propose a hybrid model combining the low-power efficiency of SNNs and high-accuracy effectiveness of ANNs for high-speed tracking tasks. We mainly make improvements on two aspects: (1) we present a hybrid data generation method and corresponding dataset benchmark, NFS-DAVIS, for frame-based and spike-based high-speed tacking scenery; (2) we develop a hybrid approach to jointly utilize the ANN-based results and SNN-based results via the TCF and attention mechanism. It shows the advanced tracking performance on PRED18 and NFS-DAVIS dataset and achieves the record-breaking tracking speed on Tianjic chips with 2083 fps speed, which is about one order of magnitude faster than ANN-based models. Our method paves a way to combine the advantage of ANN and SNN and provides a new solution for high speed tracking scenery. A future work is to extend our model to more complicated tasks and to explore more combination solutions. 
\section*{Acknowledgments}

	The work was partially supported by Tsinghua University Initiative Scientific Research Program, and Tsinghua-Foshan Innovation Special Fund (TFISF), and National Science Foundation of China (No. 61836004, 61620106010, 61621136008, 61332007, 61327902, 61876215), the National Key Research and Development Program of China (No. 2017YFA0700900), the Brain-Science Special Program of Beijing (Grant Z181100001518006), the Suzhou-Tsinghua innovation leading program 2016SZ0102, and the Project of NSF (No. 1725447, 1730309).

\bibliographystyle{unsrt}
\bibliography{ref}

\end{document}